\def\year{2022}\relax
\documentclass[letterpaper]{article}
\usepackage{aaai22}
\usepackage{times}
\usepackage{helvet}
\usepackage{courier}
\usepackage[hyphens]{url}
\usepackage{graphicx}
\usepackage{natbib}
\usepackage{caption}
\DeclareCaptionStyle{ruled}{labelfont=normalfont,labelsep=colon,strut=off}
\usepackage{algorithm}
\usepackage{algorithmic}

\usepackage{microtype}
\usepackage{times}
\usepackage{epsfig}
\usepackage{amsmath}
\usepackage{amssymb, cite}
\usepackage{hyperref}
\usepackage{makecell,multirow,diagbox}
\usepackage{colortbl,booktabs,threeparttable}
\usepackage{subfigure}

\usepackage{newfloat}
\usepackage{listings}
\floatstyle{ruled}
\newfloat{listing}{tb}{lst}{}
\floatname{listing}{Listing}

\title{Robust Data Hiding Using Inverse Gradient Attention}
\author{
    Honglei Zhang\equalcontrib\textsuperscript{\rm 1},
    Hu Wang\equalcontrib\textsuperscript{\rm 2},
    Yuanzhouhan Cao\textsuperscript{\rm 1},
    Chunhua Shen\textsuperscript{\rm 2},
    Yidong Li\textsuperscript{\rm 1}\thanks{Corresponding author.}
}
\affiliations{
    \textsuperscript{\rm 1}School of Computer and Information Technology, Beijing Jiaotong University, Beijing, China\\
    \textsuperscript{\rm 2}The University of Adelaide, Australia\\

    honglei.zhang@bjtu.edu.cn, hu.wang@adelaide.edu.au, yzhcao@bjtu.edu.cn, chunhua.shen@adelaide.edu.au, ydli@bjtu.edu.cn
}

\begin{document}

\maketitle
\begin{abstract}
Data hiding is the procedure of encoding desired information into a certain types of cover media (e.g. images) to resist potential noises for data recovery, while ensuring the embedded image has few perceptual perturbations. Recently, with the tremendous successes gained by deep neural networks in various fields, the research on data hiding with deep learning models has attracted an increasing amount of attentions. In deep data hiding models, to maximize the encoding capacity, each pixel of the cover image ought to be treated differently since they have different sensitivities w.r.t. visual quality. The neglecting to consider the sensitivity of each pixel inevitably affects the model's robustness for information hiding. In this paper, we propose a novel deep data hiding scheme with Inverse Gradient Attention (IGA), combining the idea of attention mechanism to endow different attention weights for different pixels. Equipped with the proposed modules, the model can spotlight pixels with more robustness for data hiding. Extensive experiments demonstrate that the proposed model outperforms the mainstream deep learning based data hiding methods on two prevalent datasets under multiple evaluation metrics. Besides, we further identify and discuss the connections between the proposed inverse gradient attention and high-frequency regions within images, which can serve as an informative reference to the deep data hiding research community. The codes are available at: \href{https://github.com/hongleizhang/IGA}{https://github.com/hongleizhang/IGA}.
\end{abstract}

\section{Introduction}\label{sec:intro}

The goal of data hiding is to embed a piece of general information into a cover media (e.g. images, audios or videos) without introducing significant perceptual differences from the original image. Meanwhile, the embedded messages can be robustly reconstructed under some intentional distortions~\cite{byrnes2021data,data_hiding_2020_ins}. In the paper, we focus on hiding messages into general images. According to the taxonomy of~\cite{wm_survey_2020}, data hiding generally includes digital watermarking~\cite{watermarking_2020_ins} and steganography~\cite{stegano_2022_ins}. Digital watermarking mainly focuses on protecting intellectual property with the use of hidden information, while steganography utilizes the hidden information for the purpose of encrypted transmission. Although data hiding is multipurpose, they can adopt the same means to achieve different goals. There are three key factors to measuring a data hiding model, \textit{robustness}, \textit{imperceptibility}, and \textit{capacity}~\cite{wm_survey_2015}. The robustness refers to the reliability of message reconstruction under image transformations. The imperceptibility refers to the similarity between the cover image and the encoded image. The capacity refers to the amount of information that a data hiding model can embed into a cover image. A robust data hiding system should take the above three aspects into account and satisfy the trade-off triangle relationship among them~\cite{survey2021}.

\begin{figure}[t]
\begin{center}
\includegraphics[scale=0.4]{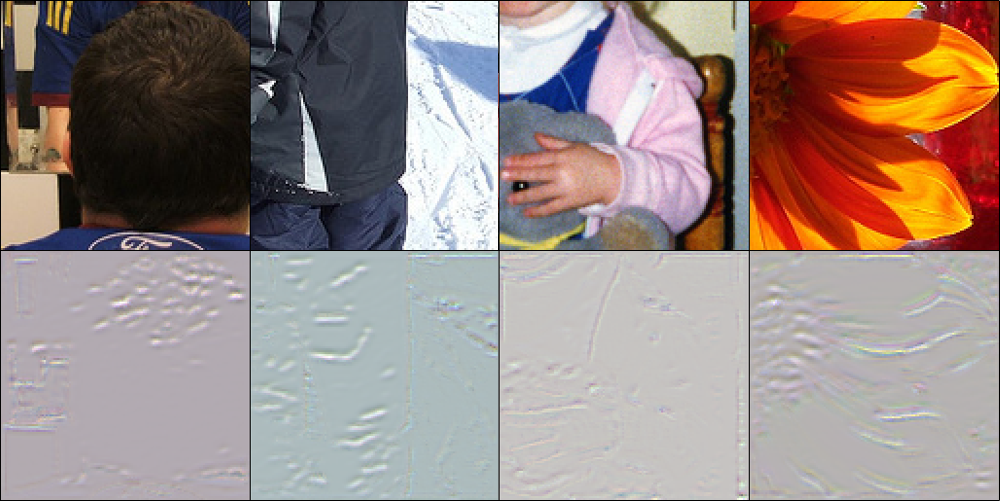}
\end{center}
\caption{Visualization of some cover images from COCO dataset and their corresponding inverse gradient attention (IGA) masks. \textbf{Top:} The cover images. \textbf{Bottom:} The IGA masks visualized by transferring them into RGB domain.
}
\label{fig:iga}
\end{figure}

The general message embedded by a robust data hiding model can survive a variety of distortions, such as JPEG compression, rotation, blurring, cropping, and quantization~\cite{eccv2018_hidden}. To achieve this goal, some traditional methods typically utilize heuristics to hide messages through texture~\cite{dh_ibm} or frequent domains~\cite{mwd_icassp_1998}. For instance, the information can be embedded in the spatial domain by substituting the least significant bits (LSB) of the pixel values~\cite{lsb2005,lsb2008}. In recent years, deep learning methods have achieved outstanding performance in the computer vision domain, as well as in data hiding, due to the massive amounts of data and rapidly increasing computation horsepower. Generally, the architecture of deep data hiding models consists of an encoder and a decoder~\cite{nips2017_deep_steg,redmark2020}. Given a piece of input message and a cover image, the encoder produces a visually indistinguishable embedded image, from which the decoder can reconstruct the original message. Zhu \textit{et al.}~\cite{eccv2018_hidden} applied a generative adversarial network (GAN) and designed a unified framework to encode a rich amount of useful information into invisible perturbations for digital watermarking and steganography. Similarly, Luo \textit{et al.}~\cite{da_cvpr20} adopted GAN as an attacking network to automatically and adaptively generate distortions. The watermarking model is with more robustness than the models trained with known distortions.

In a cover image, different pixels ought to possess different importance for data hiding to resist distortions. There are existing works exploring the attention-based models for information hiding. Nevertheless, existing deep attention-based works adopt black-box models, which are obscure for explanation. Moreover, it introduces extra learnable parameters that increase the difficulty for model optimization. Targeting at these issues, in this paper, we propose a novel data hiding method to dynamically assign different attention weights to pixels with different sensitivities for information hiding. The proposed Inverse Gradient Attention (IGA) mask is obtained by depicting the gradients of each pixel towards the message reconstruction objective. The details of the model will be illustrated in Section \ref{sec:proposed}. By doing so, it avoids introducing extra tunable parameters. Moreover, the IGA scheme is more explainable compared to the existing attention-based models since the gradient of each pixel towards the objective generally shows the pixel-wise sensitivities for message embedding, which has been well-proved by~\cite{iclr2015_fgsm}.

Fig.~\ref{fig:iga} shows intuitive examples of the inverse gradient attention mask generated by our IGA model. The inverse gradient attention mask locates the pixels that are robust for message embedding. Though it is simple, the proposed IGA scheme is able to effectively improve the performance of the data hiding model by a large margin against various image distortions.

Another challenge with data hiding is that embedding lengthy messages can easily alter the appearance and underlying statistics of the cover image~\cite{nips2017_deep_steg,capacity2014}. Therefore, to enhance the capacity while maintaining the imperceptibility of the proposed data hiding model, we further introduce a message coding module. Specifically, we adopt a symmetric encoder-decoder structure for message coding. The main contributions of the paper are listed as below:

\begin{itemize}
\item We propose a novel end-to-end deep data hiding model with inverse gradient attention mechanism (IGA), which allows the model to spotlight the pixels with more robustness for data hiding. By combining the IGA scheme, the models can hide data more accurately and robustly to resist a variety of image distortions.

\item To improve the model capacity and robustness, we propose a message coding module, as known as Message Encoder and Message Decoder in the framework, to map lengthy binary messages onto a compressed low dimensional space of real values and map back for less reconstruction disturbance.

\item Extensive experiments have been conducted on multiple prevalent datasets and instantiate them on mainstream data hiding models. Our proposed model can surpass its counterparts by a large margin and achieve promising performance under multiple settings. Furthermore, we identify and discuss the connections between the proposed inverse gradient attention with high-frequency regions within images, which can serve as an informative reference to the deep data hiding research community.
\end{itemize}

The rest of the paper is organized as follows. The related work w.r.t. traditional and deep learning based data hiding methods is comprehensively discussed in Section II. In Section III, we illustrate the proposed inverse gradient attention module and message coding module. Experimental settings and model performance evaluation are shown in Section IV. Finally, we conclude this paper and put forward the future work.

\begin{figure*}[t]
\begin{center}
\includegraphics[scale=0.7]{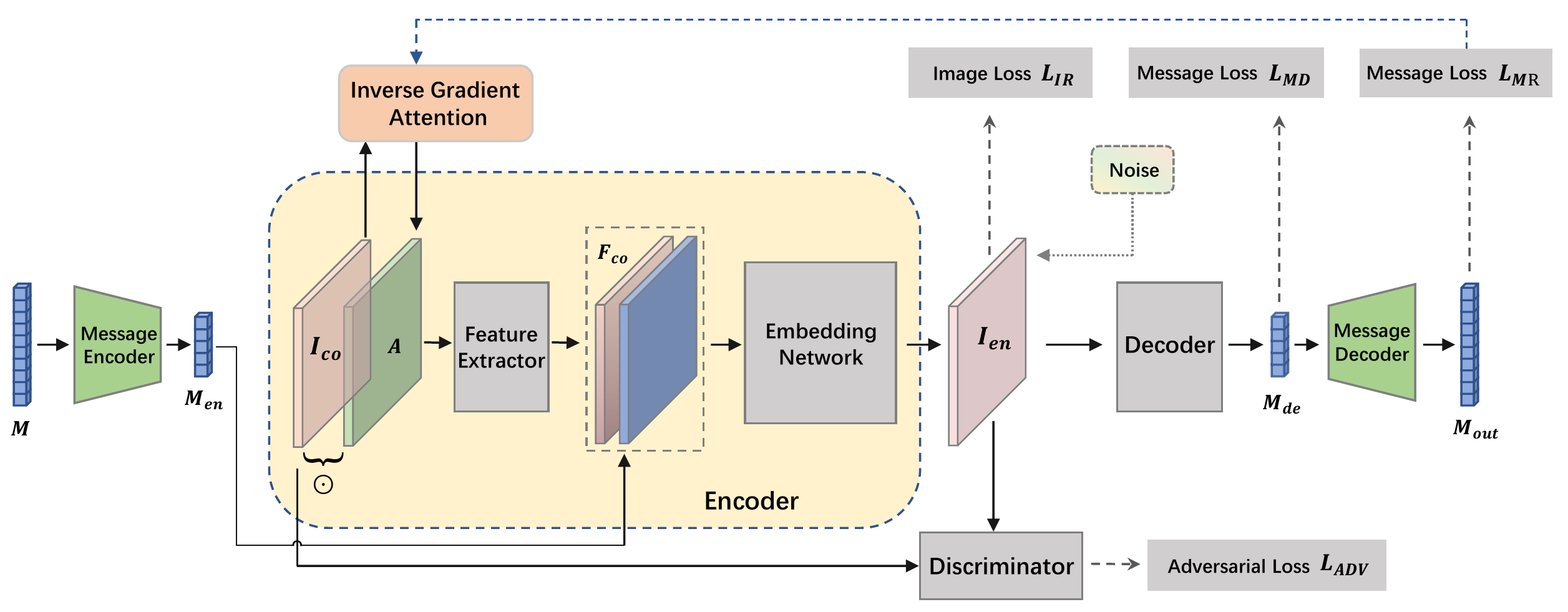}
\end{center}
\caption{The framework of the proposed Inverse Gradient Attention model. The message to be embedded is fed into the Message Encoder Network to produce a compact encoded message representation for data hiding and then map back to improve the model capacity and robustness. The inverse gradient attention mask of the cover image is computed from the pixel-wise gradients towards the message reconstruction loss through back-propagation to spotlight pixels with more robustness for data hiding.
}
\label{fig:framework}
\end{figure*}

\section{Related Work}\label{sec:relat}

Following the vein of the development of data hiding technology, we divide the related work into two categories: traditional data hiding approaches and deep learning-based data hiding approaches.

\subsection{Traditional Data Hiding Approaches}

The traditional data hiding models mainly adopt human heuristics and manual designed methods to select pixels for information embedding~\cite{dw2005survey}. According to the domains of manipulation, these models can be further divided into spatial domain data hiding \cite{lsb_2010,hugo2010,lsb_2011} and frequency domain data hiding~\cite{dct_2015,dct_2016,sd_tip2007,dwt2004,dft2008}. For spatial domain data hiding, Pevny \textit{et al.}~\cite{hugo2010} proposed HUGO algorithm to manipulate the least significant bits (LSB) of the cover image. Besides, Banitablebi \textit{et al.}~\cite{lsb_2011} proposed a robust least significant bit watermarking model to compute the structural similarity in the process of embedding and extracting watermarks. From the frequency domain perspective, a few existing algorithms changed middle frequency components of the cover image in the frequency domain \cite{sd_tip2007}, and others exploited the correlations between two Discrete Cosine Transform (DCT) coefficients of the adjacent blocks in the same position~\cite{dct_2016,dct_watermarking_2020_ins}. Reversible transformations, \textit{i.e.} the Discrete Wavelet Transform (DWT)~\cite{dwt2004} or the Discrete Fourier Transform (DFT)~\cite{dft2008}, are also widely used for information hiding.

\subsection{Deep Learning based Data Hiding Approaches}

Due to the powerful representing ability of deep neural networks, an increasing number of deep data hiding models have been proposed~\cite{survey2021,stegaStamp2020,nips2020udh,rihoop2020,lfm2019cvpr,hiwi2020pami}. For instance, Zhang \textit{et al.}\cite{nips2020udh} proposed a new universal deep hiding meta-architecture to disentangle the encoding process of the message form the cover image, and proved that the success of deep data hiding can be attributed to a frequency discrepancy between the cover image and the encoded message. Recently, the encoder-decoder framework has received more attention for data hiding since it fits the symmetrical encoding and decoding process of information embedding and extraction \cite{nips2017_deep_steg,eccv2018_hidden,attention_aaai20,redmark2020,da_cvpr20,stegAdv2017nips}. Specifically, Hayes and Ganezis \textit{et al.}~\cite{stegAdv2017nips} designed an effective deep data hiding scheme by leveraging techniques from the field of adversarial training to narrow the perceptual difference between the cover image and the encoded image~\cite{gan2014nips}. Besides, HiDDeN~\cite{eccv2018_hidden} adopted the adversarial learning mechanism and applied a set of predefined noises in a mini-batch to enhance the model robustness. It is also a unified end-to-end framework for digital watermarking and steganography. ReDMark~\cite{redmark2020} adopted a similar way by choosing one type of attacks given a certain amount of probabilities at each iteration. This technique has shown its effectiveness by improving model robustness.

Besides, Luo \textit{et al.}~\cite{da_cvpr20} proposed a distortion-agnostic model to adversarially add noises generated by an attacking network to achieve the purpose of adaptive data augmentation. It also proposed to use a channel coding mechanism to inject redundancy into encoded messages in noisy channels for model robustness improvement. Because different pixels in the cover image have different importances toward distortions, ABDH~\cite{attention_aaai20} adopted attention-based CNN model to perceive inconspicuous pixels and introduced a cycle discriminative mechanism to enhance the quality of the encoded image for information hiding.

Compared to the existing models, one of the major differences of our model is that instead of proposing a black-box neural attention module, we propose a simple yet effective model to directly generate an explainable inverse gradient attention mask to locate pixels in the cover image for robust data hiding. Moreover, to improve the model capacity and robustness, we introduce a message coding module to map the binary messages onto a compact low dimensional-space with real values.

\section{Proposed Method}\label{sec:proposed}

In this section, we first introduce the overall encoder-decoder architecture of the proposed IGA model, including message coding module and inverse gradient attention module. Then, set forth the loss functions utilized in the supervised training process, and finally list a variety of distortion operations adopted in order to increase the robustness of the data hiding model.

\subsection{Overall Architecture}
Fig.~\ref{fig:framework} depicts the overall architecture of our proposed model. Our method adds two novel components \textit{i.e.} message coding module and inverse gradient attention module, on top of the classic data hiding Encoder-Decoder framework. The message coding module is responsible for embedding the lengthy binary messages into condensed real-value ones and restoring them back, for the sake of improving model capacity and robustness. On the other hand, the inverse gradient attention module aims to spotlight pixels with more robustness for data hiding, obtained from the gradients on the cover image pixels toward message reconstruction objective. It is more interpretable compared to deep neural black-box models and it also avoids introducing more parameters. Next, we will introduce the two proposed modules in detail.

\subsection{Message Coding Module}

As mentioned above, the capacity of a data hiding model refers to the amount of embedded information. A common measurement for the amount of information embedded in cover images is bits-per-pixel (BPP). The larger the message volume, the higher the BPP. In data hiding models, the amount of information is normally set to a threshold or lower~\cite{ejasp2010_capacity} to maintain a reasonable imperceptibility. In order to enhance the capacity while maintaining the imperceptibility of our proposed data hiding model, we introduce a message coding module. The intuition behind it is, before embedding a lengthy binary message into a cover image, we can find a more condensed alternative real-value representation. The large space of real value domain will empower the learning of semantic-rich representations of message but with less bits, it thus will cause less disturbance on the premise of the same amount of information and be able to embed more information.

\begin{figure}[h]
\begin{center}
\includegraphics[scale=0.32]{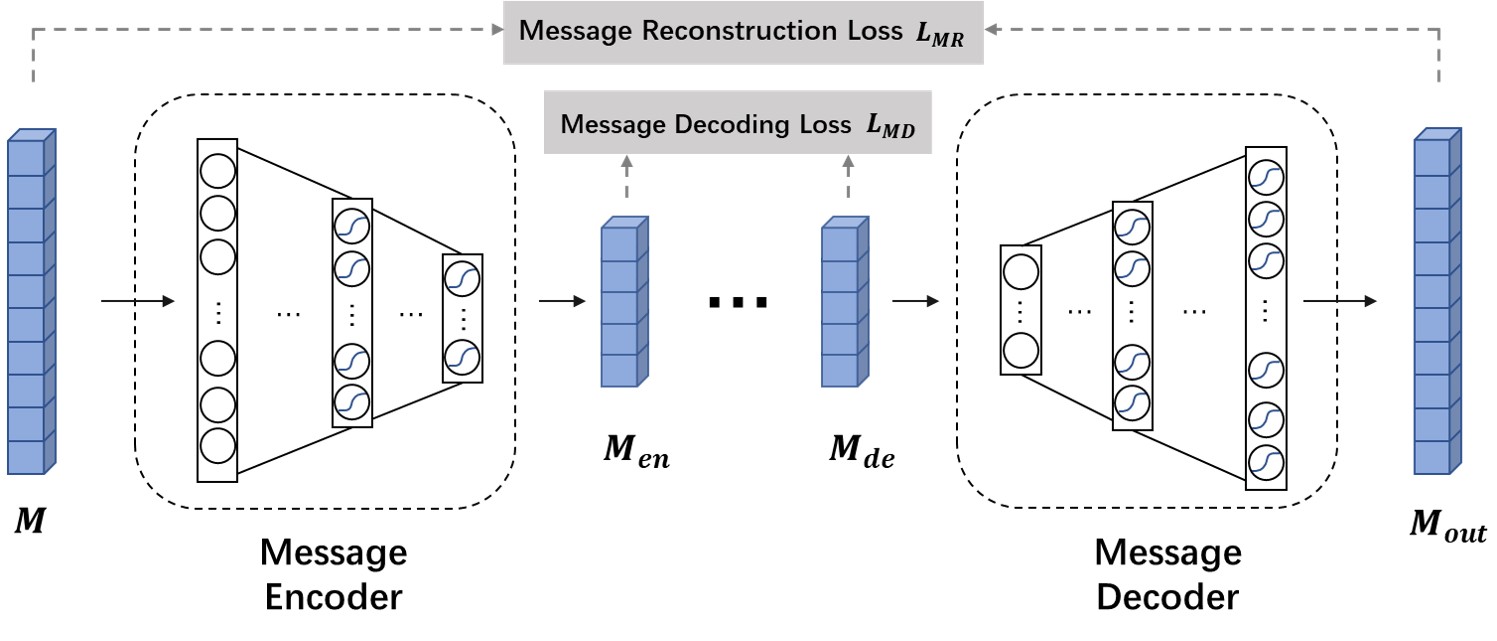}
\end{center}
\caption{Illustration of the message coding module. The message encoder maps the lengthy binary message onto a low-dimensional space to produce an encoded message with real values. Additional reconstruction supervision for messages in the real-value domain is also provided for better model optimization.
}
\label{fig:msg_pro}
\end{figure}

As theoretically proved by~\cite{uat1989}, any continuous mapping can be approximated by a neural network with one hidden layer with non-linear projection. Based on this theory, we apply multi-layer perceptrons (MLPs) for the mapping. Specifically, we adopt a symmetric structure for message coding, \textit{i.e.,} a message encoder and a message decoder, as illustrated in Fig.~\ref{fig:msg_pro}. They are all made up of MLPs with one hidden layer. The message encoder takes the lengthy binary message $\mathbf{M}\in \{0,1\}^k$ of length $k$ as input and it outputs a real-valued encoded message $\mathbf{M}_{en}\in \mathbb{R}^l$ of length $l$, where $l<k$. The embedding process is conducted through a non-linear mapping operation, and then the compact encoded message is embedded into the cover image for less reconstruction disturbance. Formally, our message encoder is defined as:

\begin{equation}
\begin{aligned}
\mathbf{z}_{0} &=\mathbf{M}^T \\
\mathbf{z}_{1} = \phi^{e}_{1}\left(\mathbf{z}_{0}\right) &=a^e_{1}\left(\mathbf{z}_{0}\mathbf{W}^e_{1}+\mathbf{b}^e_{1}\right) \\
\mathbf{z}_{2} = \phi^e_{2}\left(\mathbf{z}_{1}\right) &=a^e_{2}\left(\mathbf{z}_{1}\mathbf{W}^e_{2}+\mathbf{b}^e_{2}\right) \\
\mathbf{M}_{en} &=\sigma\left(\mathbf{z}_{2}^T\right)
\end{aligned}
\end{equation}

\noindent where $\mathbf{W}^e_{1}\in\mathbb{R}^{k\times h}$,$\mathbf{W}^e_{2}\in\mathbb{R}^{h\times l}$ denote the weight matrix, $\mathbf{b}^e_{1}\in\mathbb{R}^{h}$ and $\mathbf{b}^e_{2}\in\mathbb{R}^{l}$ denote the bias vector, $\phi^e_{x}(\cdot)$, $a^e_{x}(\cdot)$ and $\sigma(\cdot)$ denote the non-linear transformation and activation functions for the $x$-th layer's perceptron in the message encoder, respectively. For activation functions of MLP layers, one can choose sigmoid, hyperbolic tangent (Tanh), and Rectifier (ReLU), etc. Empirically, we opt for ReLU, which is more biologically plausible and theoretically proven to be non-saturated~\cite{relu}.

Then at the end of our architecture, the message decoder maps backward the decoded message $\mathbf{M}_{de}\in\mathbb{R}^{l}$ extracted from the Decoder to the recovered message $\mathbf{M}_{out}\in\mathbb{R}^{k}$. Similarly, our message decoder is defined as:

\begin{equation}
\begin{aligned}
\mathbf{h}_{0} &= \mathbf{M}_{de}^T\\
\mathbf{h}_{1}=\phi^d_{1}\left(\mathbf{h}_{0}\right) &=a^d_{1}\left(\mathbf{h}_{0}\mathbf{W}^d_{1}+\mathbf{b}^d_{1}\right) \\
\mathbf{h}_{2}=\phi^d_{2}\left(\mathbf{h}_{1}\right) &=a^d_{2}\left(\mathbf{h}_{1}\mathbf{W}^d_{2} +\mathbf{b}^d_{2}\right) \\
\mathbf{M}_{out} &=\sigma\left(\mathbf{z}_{2}^T\right)
\end{aligned}
\end{equation}

\noindent where the superscript $d$ is utilized to represent the parameters in the message decoder. The optimization of message coding is performed under the two-stage losses, \textit{i.e.} the Message Decoding Loss $\mathbf{L}_{MD}$ and the Message Reconstruction Loss $\mathbf{L}_{MR}$. Given the input message $\mathbf{M}$ and the output message $\mathbf{M}_{out}$, we can calculate a message reconstruction loss $\mathrm{L}_{MR}$. In order to obtain the better intermediate representation of raw message, we introduce a message decoding loss $\mathrm{L}_{MD}$ to relay the encoded message $\mathbf{M}_{en}$ and the decoded message $\mathbf{M}_{de}$. We emphasis here, our message coding module is different from the channel coding proposed by~\cite{da_cvpr20}. The channel coding is to produce a redundant message in noisy channels to enhance the model robustness. However, our message coding module is to reduce the dimension of the binary message onto a compact real value space to increase the model capacity and robustness.

\subsection{Inverse Gradient Attention Module}

For a data hiding model, the embedded message needs to be robustly reconstructed under image distortions. To achieve promising robustness, the pixels that are robust for message reconstruction ought to be located in the cover image, and then impose more burden of message hiding on these pixels. As indicated by FGSM~\cite{iclr2015_fgsm}, applying small but intentionally worst-case perturbations towards some pixels to the original image can result in model outputting an incorrect result with high confidence. Inspired by it, we propose a simple yet effective IGA module to locate those pixels that are robust for message reconstruction. Specifically, we first calculate a message reconstruction loss $\mathrm{L}_{MR}(\mathbf{M},\mathbf{M}_{out})$ based on the message $\mathbf{M}$ to be encoded and the reconstructed message $\mathbf{M}_{out}$. Then, an attention mask $\mathbf{A}$ is generated by calculating the inverse normalized gradients of cover image $\mathbf{I}_{co}$ toward the message reconstruction loss $\mathrm{L}_{MR}$ through back-propagation. The gradient values generally show the robustness of each pixel for message reconstruction. Formally, this process can be presented as:

\begin{equation}\label{eq:ig_attention}
    \mathbf{A} = \mathbf{T}-g(\nabla_{I_{co}}\mathrm{L}_{MR}(\mathbf{M},\mathbf{M}_{out})),
\end{equation}

\noindent where $\mathbf{T}$ represents the tensor containing all ones. $g$ denotes the general normalization function which is adopted to constrain the gradient values ranging from 0 to 1, \textit{e.g.}, the sigmoid function or min-max normalization function, as the weights of the image pixels. It is notable here that the shape of the inverse attention tensor $\mathbf{A}$ is the same as the cover image $I_{co}$. Both of them are with size $H\times W\times C$, where $H$, $W$, $C$ are the height, width and channel number of $\mathbf{A}$ and $I_{co}$. We emphasize the differences between FGSM and our proposed IGA model here. Firstly, our model is not the generation of adversarial examples, so different norms are no longer required. Besides, the tuning of $\epsilon$ adopted in~\cite{iclr2015_fgsm} is not needed, since in IGA inverse gradient attention with continuous values rather than hard sign is computed (between [0, 1]).

Intuitively, the inverse gradient attention mask highlights the pixels that are robust for message reconstruction. The lower gradient value of a pixel towards the message reconstruction objective, the smaller impact of it on recovering the message. Therefore, more weights could be allocated to such a pixel and more information could be expected to embed into it. In this case, we are able to encode messages robustly on these pixels. Particularly, we first obtain the attended image $\mathbf{I}^{A}_{co}$ by the Hadamard product of the cover image $\mathbf{I}_{co}$ with the obtained attention mask $\mathbf{A}$. The multiplication is performed in a pixel-wise manner, the attended image  is therefore obtained. Meanwhile, the encoded message $\mathbf{M}_{en}$ is expanded into the massage matrix with the shape of $H\times W \times l$ to align with the attended image, where $H$, $W$, $l$ are the height, width and channel number of the message matrix. Then, the attended image is fed into the feature extractor $\mathcal{E}$ with the message matrix to produce intermediate feature maps $\mathbf{F}_{co}$:
\begin{equation}\label{eq:ig_ico_attention}
    \mathbf{F}_{co} = \mathcal{E}(\mathbf{A} \odot \mathbf{I}_{co}) \oplus \mathbf{M}_{en},
\end{equation}
where $\mathcal{E}$ is the feature extractor. $\mathbf{M}_{en}$ is generated by the message encoder in the message coding module. $\odot$ denotes the Hadamard product and $\oplus$ represents the concatenation operation. The intermediate feature $\mathbf{F}_{co}$ is fed into the embedding network to generate the encoded image $\mathbf{I}_{en}$. After the decoder network produces a reconstructed decoded message $\mathbf{M}_{de}$, $\mathbf{M}_{de}$ is further fed into the message decoder network to produce the final recovered message $\mathbf{M}_{out}$. The training process of the framework is optimized under the supervision of the following four objectives: Image Reconstruction objective $\mathrm{L}_{IR}$, Message Decoding objective $\mathrm{L}_{MD}$ and Message Reconstruction objective $\mathrm{L}_{MR}$, and Generative Adversarial objective $\mathrm{L}_{ADV}$. In the standard training process, the gradient descent optimization facilitates the attention to converge to a few most sensitive parts of the cover image, while it ignores the other less sensitive parts. The IGA training procedure iteratively converts the original gradient tensor into the inverse attention tensor. It forces the network to perceive less sensitive pixels for message reconstruction and in turn indicates how much weights should be assigned to the corresponding pixels, so as to achieve the purpose of embedding information robustly.

\subsection{Objectives}

We apply four objective functions to train our data hiding model: two message objectives to ensure the model robustness, an image reconstruction loss and an adversarial loss function to ensure the model imperceptibility. The Mean Square Error (MSE) is adopted for two message objective measurements:
\begin{equation}\label{eq:lml}
\mathrm{L}_{MR} = \lambda_{MR}\frac{1}{k} \sum_{p} ( \mathbf{M}(p)-\mathbf{M}_{out}(p) )^2,
\end{equation}

\noindent and
\begin{equation}\label{eq:lmd}
\mathrm{L}_{MD} = \lambda_{MD}\frac{1}{l} \sum_{p} ( \mathbf{M}_{en}(p)-\mathbf{M}_{de}(p) )^2,
\end{equation}

\noindent where $p$ denotes each element in the message. $\lambda_{MR}$ and $\lambda_{MD}$ represent the weight of message reconstruction loss and decoding loss, respectively. $\mathrm{L}_{MR} $ enforces the reconstructed message to be close to the original input, and  $\mathrm{L}_{MD} $ enforces the decoded message to be close to the encoded message.

We also adopt MSE as our image reconstruction objective to enforce the encoded image to be close to the cover image:
\begin{equation}\label{eq:lr}
\mathrm{L}_{IR} = \lambda_I\frac{1}{N} \sum_{i,j} ( \mathbf{I}_{co}(i,j) - \mathbf{I}_{en}(i,j) )^2,
\end{equation}
where $i$ and $j$ represent the pixel location and $N$ is the total number of pixels. Besides, $\lambda_I$ denotes the weight factor of the message reconstruction loss.

The image reconstruction loss focuses on pixel-level differences, so it may result in low human perceptual qualities. In order to further enforce the imperceptibility of our model, follow \cite{eccv2018_hidden}, we apply the generative adversarial learning scheme by introducing a discriminator to increase the perception similarity between the encoded image and the original one.
Specifically, we treat the encoder as a generator to produce the encoded image similar to the cover image that attempts to confuse the discriminator. The discriminator is to recognize the encoded images from the original ones. The objective of our generative adversarial learning is:

\begin{small}
\begin{equation}\label{eq:l_gan}
\mathrm{L}_{ADV}=\begin{cases}
\max \limits_{D} \quad \lambda_D \mathbb{E}_{\mathbf{x}\in\zeta}[\log(D(\mathbf{x}))+\log(1-D(G(\mathbf{x})))], \\
\\
\min \limits_{G} \quad \lambda_G \mathbb{E}_{\mathbf{x}\in\zeta}[\log(1-D(G(\mathbf{x})))],\\
\end{cases}
\end{equation}
\end{small}

\noindent where $\mathbf{x}$ is the input cover image and $\zeta$ represents its distributions, and $G$ denotes the generator and $D$ is the discriminator. The variables $\lambda_{D}$ and $\lambda_{G}$ denote the weight factors of the maximization and minimization process. The maximization step aims to find the most determined discriminator model against a given generator model. The minimization process aims to seek the strongest generator model against a given discriminator model. These two optimizations need to be solved alternately to reach the equilibrium point. When the whole process achieves Nash Equilibrium where the discriminator returns the classification probability 0.5 for each pair of encoded image and cover image~\cite{gan2014nips}, the image with embedded information is almost completely indistinguishable from the original image.

\begin{table}[t]
\renewcommand\arraystretch{1.4}
\footnotesize
\caption{The detailed descriptions of distortions adopted in our experiments.
}
\begin{center}
  \begin{tabular}{c|c|m{3.6cm}}
    \toprule
    \textbf{Noises} & \textbf{Noisy images} $\mathbf{I}_{no}$ & \textbf{Descriptions} \\
    \hline
        \textit{Identity} & $\mathbf{I}_{en}$ & It makes $\mathbf{I}_{en}$ unchanged. \\
    \hline
        \textit{Crop} & $\mathcal{C}(\mathbf{I}_{en})$ & It randomly produces a sub square area with $H^{*}\times W^{*}$ from the encoded image $\mathbf{I}_{en}$ with the ratio $p=\frac{H^{*}\times W^{*}}{H\times W}\in (0, 1)$.   \\
    \hline
        \textit{Cropout} & $p_c\mathcal{C}(\mathbf{I}_{en})+(1-p_c)\mathbf{I}_{co}$ & It combines the pixels from the cropped image $\mathcal{C}(\mathbf{I}_{en})$ with the percentage $p_c\in (0, 1)$ and the rest from the cover image $\mathbf{I}_{co}$.   \\
    \hline
        \textit{Dropout} & $p_d\mathbf{I}_{en}+(1-p_d)\mathbf{I}_{co}$ & It combines the pixels from the encoded image $\mathbf{I}_{en}$ with the percentage $p_d\in (0, 1)$ and the rest from the cover image $\mathbf{I}_{co}$. \\
    \hline
        \textit{Resize} & $\mathcal{R}(\mathbf{I}_{en})$ & It is obtained from the original encoded image $\mathbf{I}_{en}$ with a certain zoom factor $z\in (0, 1)$.  \\
    \hline
        \textit{Jpeg} & $\mathcal{J}(\mathbf{I}_{en})$ & It applies JPEG compression to the encoded image $\mathbf{I}_{en}$ with  quality factor $q\in (0, 100)$.\\
    \bottomrule
  \end{tabular}
  \end{center}
  \label{tab:distortion}
\end{table}

Note that the setting and purpose of our generative adversarial learning are different from \cite{da_cvpr20}. The image generator in \cite{da_cvpr20} is used to adaptively generate image distortions for the sake of resisting unknown noises, while our work is to generate realistic encoded images through adversarial training between the discriminator and generator, and we focus on evaluating the performance of models on specific distortions as described detailedly in Table~\ref{tab:distortion}, such as \textit{Crop}, \textit{Cropout}, \textit{Resize}, \textit{Dropout} and \textit{Jpeg compression} to the encoded image.

\section{Experiments}\label{sec:exp}

The experiments contain four parts: in the first two parts, we will compare our approach with other strong data hiding baseline models in three aspects, i.e., \textit{robustness}, \textit{imperceptibility} and \textit{capacity}; the ablation study is further presented to verify the contributions of each component within our framework. Finally, we discuss the relations between the proposed IGA and high-frequency image regions to offer more insights.

\begin{table*}[htb]
\renewcommand\arraystretch{1.5}
\scriptsize
\caption{Comparisons of the proposed IGA model with HiDDeN, ABDH and Distortion-Agnostic (DA) model on the COCO and the DIV2K datasets. The best result for each column is bolded.}
\begin{center}
  \begin{tabular}{ccccccccc|ccccccc}
    \toprule
    \multirow{2}*{\textbf{Methods}}  && \multicolumn{7}{c}{COCO}& \multicolumn{7}{c}{DIV2K}\\
    \cline{2-16}
     & $k$ & \textit{Identity} & \textit{Crop} & \textit{Cropout} & \textit{Dropout} & \textit{Resize} & \textit{Jpeg} & \textit{CN} & \textit{Identity} & \textit{Crop}& \textit{Cropout} & \textit{Dropout} & \textit{Resize} & \textit{Jpeg} & \textit{CN} \\
    \midrule
    \multirow{3}*{HiDDeN} &30 & 98.10 & 80.34 & 75.96 & 76.89 & \textbf{82.72} & 84.09 & 76.30& 73.72 & 68.24 &60.92 & 63.78& 66.28&66.37 & 58.05  \\
                          &64 & 79.82 & 72.52 & 63.20 & 68.53 & 69.35 & 68.85 & 65.46 & 70.45 & 51.60 & 51.09& 52.40&50.81 & 50.99 & 52.35 \\
                          &90 & 73.56 & 65.46 & 60.20 & 61.49 & 63.03 & 63.21 & 62.07& 58.04 & 50.63 &51.50 &50.22 & 51.40& 51.16& 51.84 \\

    \hline
    \multirow{3}*{ABDH} &30 & 98.70 & 85.16 & 74.82 & 75.31 & 80.23 & 82.62 & 74.01& \textbf{81.09} &62.24 & 59.71 &58.72 & 60.83& 63.44& 63.58 \\
                      &64 & 85.93 & 67.14 & 59.96 & 62.62 & 62.02 & 61.74 & 59.14& 64.58 & 51.69& 52.21 & 53.32& 55.20& 52.72& 50.78 \\
                      &90 & 72.22 & 52.65 & 51.99 & 52.88 & 52.95 & 53.04 & 52.25& 56.44 & 51.48& 51.67 &51.28 & 51.15& 51.19& 50.23 \\
    \hline
    \multirow{3}*{DA} &30 & 99.50 & 81.15 & 78.58 & 77.13 & 81.72 & 82.83 & 75.73& 78.80 &77.32 & \textbf{77.11} &74.55 & 71.01& 82.35& 63.85 \\
                      &64 & 77.18 & 68.68 & 65.82 & \textbf{73.00} & 62.54 & \textbf{73.58} & 64.90& 62.01 & 62.65& 61.79 & \textbf{71.09}& 58.91& 71.26& 53.31 \\
                      &90 & 70.82 & 63.51 & 61.08 & 64.28 & 62.62 & 67.00 & 63.21& 57.88 & 58.12& 54.18 &57.32 & 55.68& 62.17& 53.05 \\
     \hline
    \multirow{3}*{IGA (Ours)} &30 & \textbf{99.96} & \textbf{86.88} & \textbf{79.33} & \textbf{77.51} & 81.44 & \textbf{87.35} & \textbf{82.30} & 79.94 & \textbf{77.39} & 60.93 & \textbf{76.63} & \textbf{72.19}& \textbf{82.90} & \textbf{64.14} \\
                       &64 & \textbf{94.34} & \textbf{73.34} & \textbf{66.82} & 70.23 & \textbf{69.41} & 72.07 & \textbf{78.38} & \textbf{77.61} & \textbf{63.11}& \textbf{62.00} &70.14 &\textbf{59.64} &\textbf{72.31} & \textbf{57.03} \\
                       &90 & \textbf{74.62} & \textbf{69.45} & \textbf{62.52} & \textbf{64.96} & \textbf{65.78} & \textbf{68.78} & \textbf{71.47} & \textbf{62.26} &\textbf{59.10} & \textbf{55.31} & \textbf{60.10}& \textbf{56.48} & \textbf{63.21} & \textbf{55.32} \\
    \bottomrule
  \end{tabular}
  \end{center}
  \label{tab:hidden}
\end{table*}

\subsection{Experimental Settings}

\textbf{Datasets.} In order to verify the effectiveness of the proposed model, we adopt two real-world datasets for model training and evaluation, the COCO dataset \cite{coco} and the DIV2K dataset \cite{div2k}. For the COCO dataset, 10,000 images are collected for training, and evaluation is performed on the other 1000 unseen images. For the DIV2K dataset, we use 800 images for training and 100 images for evaluation. For each image in the two datasets, a corresponding string message is generated with a fixed length $k$.

\textbf{Evaluation Metrics.} To thoroughly evaluate the performance of the proposed model and other data hiding models, we apply a series of evaluation metrics. For model \textit{robustness}, the bit prediction accuracy (BPA) is utilized to evaluate the ability of data hiding model to withstand image distortions. It is defined as the fraction of the common bits between their input message $M_{in}$ and the reconstructed message $M_{out}$. Formally, the BPA can be expressed as:

\begin{equation}\label{eq:bpa}
BPA(\mathbf{M}_{in},\mathbf{M}_{out}) = \frac{\sum_p{\mathbb{I}(\mathbf{M}_{in}(p)=\mathbf{M}_{out}(p))}}{k},
\end{equation}

\noindent where $\mathbb{I}(\cdot)$ denotes an indicator that returns 1 if the given statement is true and 0 otherwise. $k$ represents the length of the embedded message. For \textit{imperceptibility}, we adopt the peak signal-to-noise ratio (PSNR) for evaluation, and the value is calculated by the following equation:

\begin{equation}\label{eq:psnr}
PSNR(\mathbf{x}, \mathbf{y})=10 \log _{10}\frac{\left(max_{I}\right)^{2}}{mse(\mathbf{x}, \mathbf{y})},
\end{equation}

\noindent where $\mathbf{x}$ and $\mathbf{y}$ represent the cover image and encoded image, respectively. The $max_{I}$ denotes the maximum pixel value of images. $mse(\mathbf{x}, \mathbf{y})$ represents the mean squared error between images $\mathbf{x}$ and $\mathbf{y}$. In addition to the quantitative metric of PSNR w.r.t. imperceptibility, we also present visualization results of the encoded images produced by our IGA model and other comparing methods. For the model \textit{capacity}, we apply the Reed-Solomon bits-per-pixel (RS-BPP) \cite{stegagan19} as the metric representing the average number of bits that can be reliably transmitted in an image. According to the implementation of the original paper, the definition is:

\begin{equation}\label{eq:rsbpp}
RS{-}BPP(k, p) = k \times (2p-1),
\end{equation}

\noindent where $k$ represents the message length, and $p$ denotes the probability of a given model can correctly decode one bit of the embedded message. The higher the value of RS-BPP, the greater the capacity of the model. From the above equation Eq.~\ref{eq:rsbpp} we can observe that if the BPA is 0.5, the value of RS-BPP is 0. It is also in line with our intuition that if the prediction probability is around 0.5, the result is nearly random guess, and apparently, the capacity is close to null.

Worth to be noted here, data hiding schemes are characterized by three requirements: robustness against distortions in the transmission channel, capacity regarding the embedded payload of the model, and imperceptibility in terms of similarity between the cover image and the encoded image. The aforementioned metrics (in Equation \ref{eq:bpa}, \ref{eq:psnr} and \ref{eq:rsbpp}) are trade-offs for model robustness, capacity, and imperceptibility. The model with higher capacity often incurs lower imperceptibility. For data hiding tasks, including digital watermarking and steganography, we focus on the model robustness to survive from distortions and capacity of embedded payload, under the premise of ensuring imperceptibility.

\textbf{Comparing Models.} To evaluate the effectiveness of our framework in multiple paradigms, we compare a variety of mainstream data hiding models. A brief introduction to these methods are listed below:

\begin{itemize}
\item HiDDeN \cite{eccv2018_hidden} is a unified end-to-end CNN model for digital watermarking and steganography.
\item SteganoGAN \cite{stegagan19} introduces residual learning into the data hiding process and is able to embeds messages with different channel depths.
\item Distortion-Agnostic \cite{da_cvpr20} can resist unknown image distortions by adaptively adding noises through adversarial learning.
\item ABDH \cite{attention_aaai20} proposes to learn an attention mask through neural networks to locate the inconspicuous areas of cover images for message embedding\footnote{For fair comparisons, we instantiate ABDH to convert the message strings into message matrices as the secret image for embedding binary messages into the cover image.}.
\end{itemize}

\textbf{Implementation Details.} In our implementation, images are resized to $128\times128$ for HiDDeN, Distortion-Agnostic and ABDH models; while for SteganoGAN model, images are resized to $400\times400$. To keep a fair comparison, we adopt the exactly same settings with the comparing methods\footnote{Across all tables, the IGA$^{*}$ model represents the extended version of the proposed IGA model by equipping the proposed two components on SteganoGAN with different message channels.}. The \textit{Combined Noises} (CN) is adopted for all model training. The noise set includes $\{\textit{Crop},\textit{Cropout},\textit{Resize}, \textit{Dropout}\,\, and\,\, \textit{Jpeg compression}\}$. At the validation phase, we evaluate the well-trained model by testing on each noise to verify the robustness of the model against various distortions.

For the proposed IGA model, the weight factors $\lambda_{MR}$, $\lambda_{MD}$, $\lambda_I$, $\lambda_D$ and $\lambda_G$ are set to 1.0, 0.001, 0.7, 1.0 and 0.001, respectively. In the training phase, we adopted the Adam as the optimizer \cite{adam} with default hyper-parameters. The batch size is set to 32. The Message Encoder and Message Decoder are made up of fully connected networks with non-linear transformations. The Feature Extraction Network, Embedding Network, Decoder Network and Discriminator Network are all composed of convolution layers.

\begin{table}[t]
\renewcommand\arraystretch{1.5}
\scriptsize
  \caption{Comparisons of the proposed IGA model with the SteganoGAN model on COCO and DIV2K datasets. The best result for each column is bolded.}
\begin{center}
  \begin{tabular}{cccc|cc}
    \toprule
    \multirow{2}*{\textbf{Methods}}  && \multicolumn{2}{c}{COCO}& \multicolumn{2}{c}{DIV2K}\\
    \cline{2-6}
     & $D$ & \textit{Identity}  & \textit{CN}  & \textit{Identity} & \textit{CN}   \\
    \midrule
    \multirow{5}*{SteganoGAN} &1 & 97.91 & 62.38  &98.29  & 61.38 \\
                      &2 & 96.02 & 62.63  &96.53  & 58.11 \\
                      &3 & 86.19 & 56.64  &89.10  & 55.86 \\
                      &4 & 76.10 & 53.39  &78.07  & 53.03 \\
                      &5 & 70.98 & 53.41  &71.31  & 52.65 \\
    \hline
    \multirow{5}*{IGA$^{*}$ (Ours)} &1 & \textbf{99.67} & \textbf{68.70}  &\textbf{99.39}  & \textbf{67.65} \\
                       &2 & \textbf{99.07} & \textbf{65.04}  &\textbf{98.62}  & \textbf{58.65} \\
                       &3 & \textbf{95.26} & \textbf{59.05}  &\textbf{95.64}  & \textbf{57.92} \\
                       &4 & \textbf{84.56} & \textbf{55.78}  &\textbf{84.28}  & \textbf{57.46} \\
                       &5 & \textbf{77.78} & \textbf{55.02}  &\textbf{77.97}  & \textbf{56.59} \\
    \bottomrule
  \end{tabular}
  \end{center}
  \label{tab:SteganoGAN}
\end{table}

\begin{table}[t]
\renewcommand\arraystretch{1.5}
\scriptsize
\caption{Quantitative comparisons of encoded image quality varying different message channels across two datasets, where $d$ denotes the number of message channels.}
\begin{center}
  \begin{tabular}{cccc|cc}
    \toprule
    \multirow{2}*{\textbf{Methods}}  && \multicolumn{2}{c}{COCO}& \multicolumn{2}{c}{DIV2K}\\
    \cline{2-6}
     & $D$ & \textit{Identity}  & \textit{CN}  & \textit{Identity} & \textit{CN}   \\
    \midrule
    \multirow{5}*{SteganoGAN} &1 & 27.10 & 30.10 & 39.03 & 43.01 \\
                      &2 & 26.43 & 31.87 & 37.56 & 46.02 \\
                      &3 & 27.16 & 32.21 & 36.98 & 46.02  \\
                      &4 & 27.21 & 31.7 & \textbf{38.23} & 46.02 \\
                      &5 & 26.88 & 30.8 & 40 & 46.02 \\

    \hline
    \multirow{5}*{IGA$^{*}$ (Ours)}&1 & \textbf{31.40} & \textbf{32.80} & \textbf{43.01} & \underline{43.01} \\
                      &2 & \textbf{31.10} & \textbf{32.04} & \textbf{40} & \underline{46.02} \\
                      &3 & \textbf{30.45} & \textbf{32.40} & \textbf{46} & \underline{46.02} \\
                      &4 & \textbf{30.96} & \textbf{32.40} & 38 & 46.02 \\
                      &5 & \textbf{30.83} & \textbf{32.59} & \underline{40} & \underline{46.02} \\

    \bottomrule
  \end{tabular}
  \end{center}
  \label{tab:imp}
\end{table}

\subsection{Quantitative Analysis on Robustness}

In this section, we evaluate the model robustness through the bit prediction accuracy and compare the proposed model with other data hiding methods. All experiments are conducted on the COCO and the DIV2K datasets. For HiDDeN, ABDH and Distortion-Agnostic model, the embedded message is a one-dimensional binary string $\mathbf{M}\in\{0,1\}^k$, we thus compare our method with the three methods and illustrate the results in Table~\ref{tab:hidden}. All models were trained and evaluated with various message lengths and image distortions. We adopt message lengths $k=30, 64, 90$, and 6 image distortions across all experiments in the table. \textit{Identity} refers to no distortion and \textit{CN} represents the combination of noises \textit{Crop}, \textit{Cropout}, \textit{Dropout}, \textit{Resize} and \textit{Jpeg} compression during training and evaluation. As for the SteganoGAN model, the embedded message is a binary tensor $\mathbf{M}\in\{0,1\}^{D\times H \times W}$, we compare the proposed method with the SteganoGAN model in Table~\ref{tab:SteganoGAN}. All models are trained and evaluated with various message lengths and image distortions. We adopt 5 message channels $D=1, 2, 3, 4, 5$ and 1 image distortion. \textit{Identity} refers to no distortion and \textit{CN} refers to the combined noise of \textit{Crop}, \textit{Cropout}, \textit{Dropout}, \textit{Resize} and \textit{Jpeg} compression during training and evaluation.

\begin{figure*}[t]
\begin{center}
\includegraphics[scale=0.63]{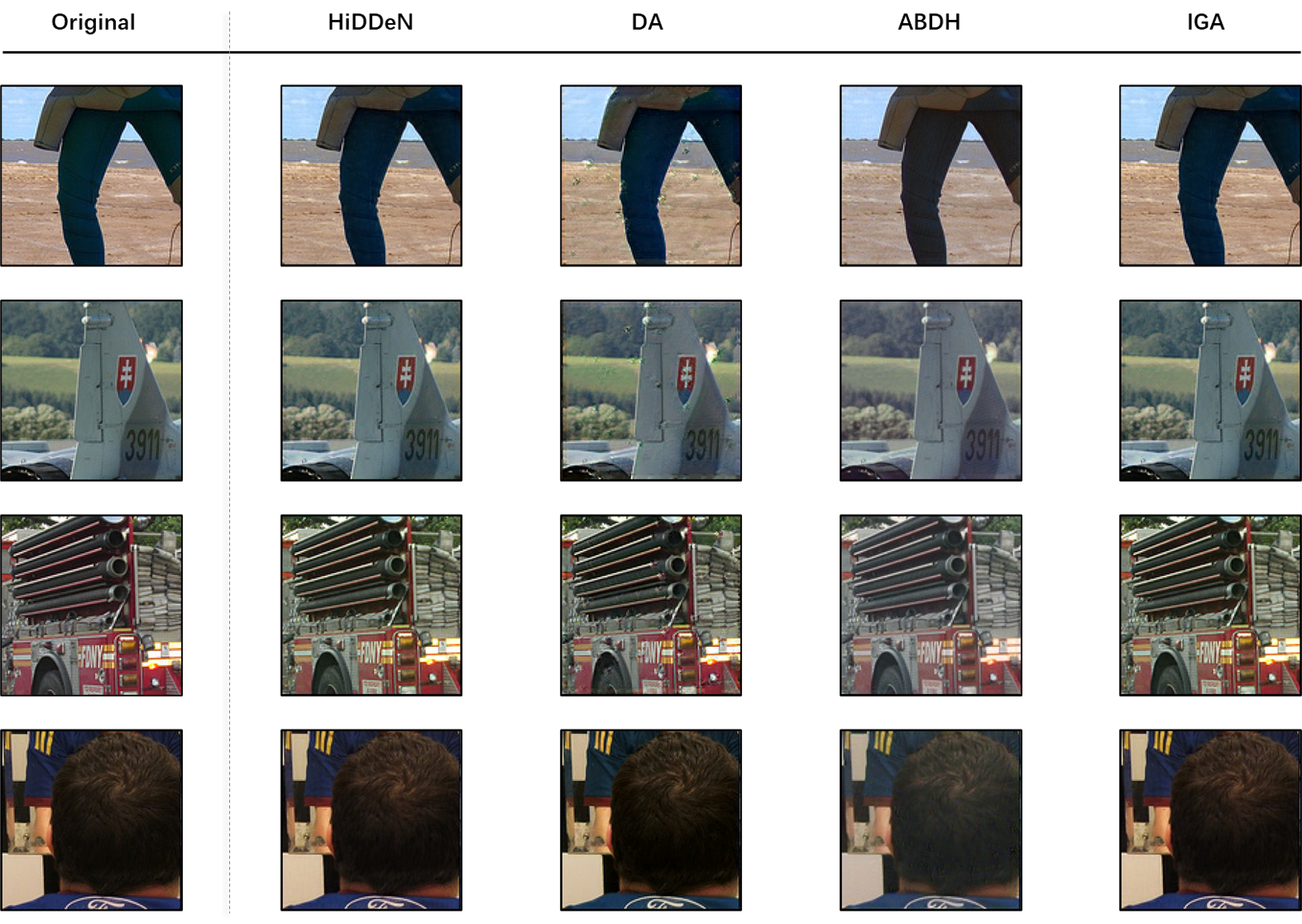}
\end{center}
\caption{Multiple samples of original and encoded images from the COCO dataset for data hiding. The first column presents the original cover images. The remaining columns depict the visualization results of the embedded images generated by different models.
}
\label{fig:imp}
\end{figure*}

\begin{figure}[t]
\begin{center}
\includegraphics[scale=0.52]{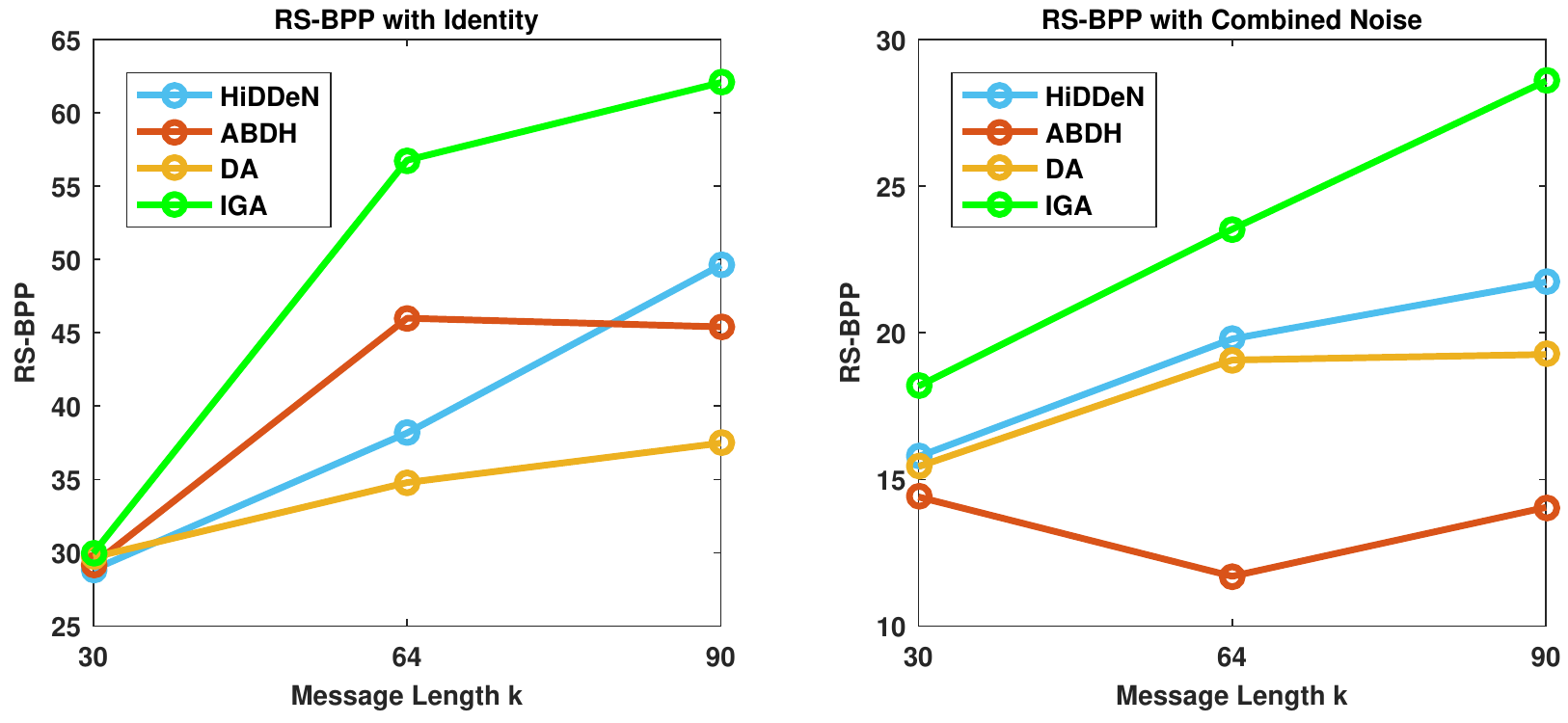}
\end{center}
\caption{Capacity comparison of the proposed model with HiDDeN, ABDH and DA (Distortion-Agnostic) models on COCO dataset under Identity and Combined Noises settings.}
\label{fig:rsbpp_coco}
\end{figure}

\begin{figure}[t]
\begin{center}
\includegraphics[scale=0.52]{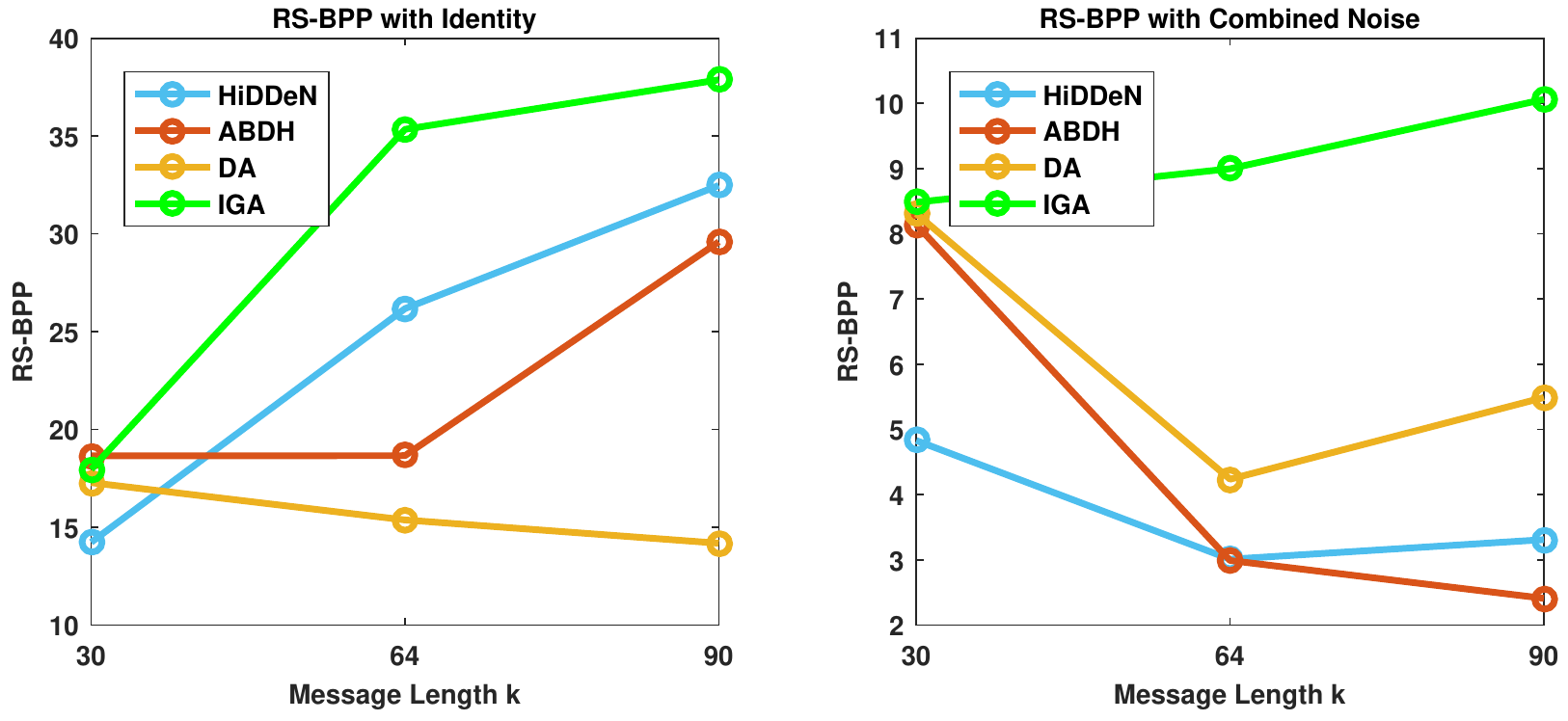}
\end{center}
\caption{Capacity comparison of the IGA model with comparing models on DIV2K dataset under Identity and Combined Noises settings.}
\label{fig:rsbpp_div2k}
\end{figure}

We can easily perceive from Table~\ref{tab:hidden}, our method outperforms the three methods in the majority of the settings. The proposed IGA model surpasses HiDDeN model by 1.86\%, 14.52\%, 7.06\% on COCO dataset and 6.22\%, 7.16\%, 3.46\% on DIV2K dataset under \textit{Identity} and message length \{30, 64, 90\} settings, respectively. When the message length is 30 or 64, only under a few image distortions that our results are slightly lower than the comparing models. Under the setting of the message length equals to 90, our model is significantly better in all image distortions. The proposed model surpasses HiDDeN model by 7.06\%, 3.46\% under \textit{Identity} and 2.25\%, 5.3\% with \textit{Combined Noises} on COCO dataset and DIV2K dataset, respectively. This phenomenon shows that our method is able to embed rich information robustly under various distortions. It is worth mentioning that our IGA model is consistently superior compared to the ABDH model under most cases. It reflects that our proposed attention mechanism generated through normalized inverse gradients is more effective and suitable for the data hiding task than the traditional black-box attention model.

Table~\ref{tab:SteganoGAN} shows the comparisons of our model with the SteganoGAN model. Note that the SteganoGAN model embeds a message tensor with different channels $D$ instead of a message string. Thus, we equip the proposed two components on SteganoGAN, as the IGA$^{*}$ model shown in the table. From the results, we can see that our method achieves the best performance with different number of message embedding channels under various noises on both datasets. Moreover, it is worth mentioning that our method has better accuracy with message channel $D=5$ than the SteganoGAN model with message channel $D=4$ on the COCO dataset. Also, the proposed model achieves comparable results on the DIV2K dataset. It indicates that our method can embed more information than the other models when the accuracy is similar. It also reflects the proposed IGA model plays a positive role in improving model capacity.

\subsection{Qualitative Analysis on Imperceptibility and Capacity}

Besides the model robustness, the quality of embedded image is also critical for data hiding task. The robustness and imperceptibility are a trade-off. If more messages are intended to be encoded, it will inevitably decline the quality of the encoded image. We conducted the experiments w.r.t. the quantitative comparisons of imperceptibility under the settings of \textit{Combined Noises} and the message length $k=30$. The results of the proposed method are comparable with HiDDeN, ABDH and Distortion-Agnostic models. Fig.~\ref{fig:imp} presents the visualization results of the proposed IGA model and the counterparts. The figure generally shows our method not only has exceptional robustness, but also takes the imperceptibility of the embedded image into account. It can be observed that the encoded images generated by our method and the cover images are visually similar. In addition, the general encoded image PSNR comparison of the proposed IGA model with the SteganoGAN model is provided in Table~\ref{tab:imp}. In the table, the higher the PSNR values, the better quality of the encoded image. Besides, the first place of each column is bolded and the underline symbol `$\_$' indicates tied performances to the best comparing models. From the table, our method can generate images with best quality under both \textit{Identity} and \textit{Combined Noises} settings on the COCO dataset. Moreover, the performance of our method is comparable in most cases on the DIV2K dataset. This experiment proves the proposed IGA model can encode messages into images resisting various noises under the premise of imperceptibility.

To measure the capacity of data hiding models, Fig.~\ref{fig:rsbpp_coco} and Fig.~\ref{fig:rsbpp_div2k} present the RS-BPP values of our model compared with HiDDeN, ABDH and Distortion-Agnostic methods on both COCO and DIV2K datasets. We can observe the trend that the capacity of the model gradually increases with length increment of the embedded message. Moreover, the capacity performance of our method is higher than the comparing methods by a large margin, whether it is under the case of \textit{Identity} or \textit{Combined Noises} settings. In light of this observation, our proposed model is verified to improve the model capacity significantly for lengthy message embedding.

\subsection{Ablation Study}

In the section, we evaluate the performance contribution of the proposed message coding module and inverse gradient attention module in Table~\ref{tab:ablation}. We conduct experiments on both COCO and DIV2K datasets with the message length $k=90$ and compressed message length $l=30$ after message encoding. For the \textit{Basic} model, it is based on the basic Encoder-Decoder framework with generative adversarial loss, and the \textit{Both} model is built on the basis of the \textit{Basic} model by equipping with inverse gradient attention module and message coding module. Besides, the \textit{w Att.} model is only equipped with the inverse gradient attention module, and the \textit{w MC.} is only equipped with the message coding module. From Table~\ref{tab:ablation}, it is noticeable that both the message coding module and the inverse gradient attention module make positive impacts on the model performance. The model performance improvement mainly comes from the inverse gradient attention module. Message coding works as an auxiliary module, and also shows its effectiveness under most circumstances. Only in one case, the performance of message coding module is slightly lower under the CN setting. We hypothesize that the combination of noises complicates the features by encoding messages from real value domain incurring difficulties for model optimization under this circumstance. Besides, a feedforward network with a single layer is sufficient to represent any function, but the layer may be infeasibly large and may fail to generalize well. In many cases, utilizing deeper models can reduce the number of parameters required to express the desired function and can decrease the amount of generalization error. Thus, we will explore deeper neural networks as our message coding module in the future to accelerate model convergence and boost the capability of generalization.

\begin{table}
\renewcommand\arraystretch{1.5}
\footnotesize
  \caption{Ablation study results on the COCO and the DIV2K datasets. The best result for each column is bolded.}
\begin{center}
  \begin{tabular}{ccc|cc}
    \toprule
    \multirow{2}*{\textbf{Methods}}  & \multicolumn{2}{c}{COCO}& \multicolumn{2}{c}{DIV2K}\\
    \cline{2-5}
     & \textit{Identity}  & \textit{CN}  & \textit{Identity} & \textit{CN}   \\
    \midrule
    \textit{Basic}  & 73.56 & 62.07  &58.04  & 51.84 \\

    \hline
    \textit{\textit{w MC.}}  & 73.89 & 65.93  & 61.21  & 51.67 \\

    \hline
    \textit{\textit{w Att.}}  & 74.47 & \textbf{73.88}  & 61.47  &53.59 \\
    \hline
    \textit{Both}  & \textbf{74.62} & 71.47  & \textbf{62.26} & \textbf{55.32} \\

    \bottomrule
  \end{tabular}
  \end{center}
  \label{tab:ablation}
\end{table}

\subsection{Discussion}\label{sec:discussion}

\begin{figure}[t]
\begin{center}
\includegraphics[scale=0.4]{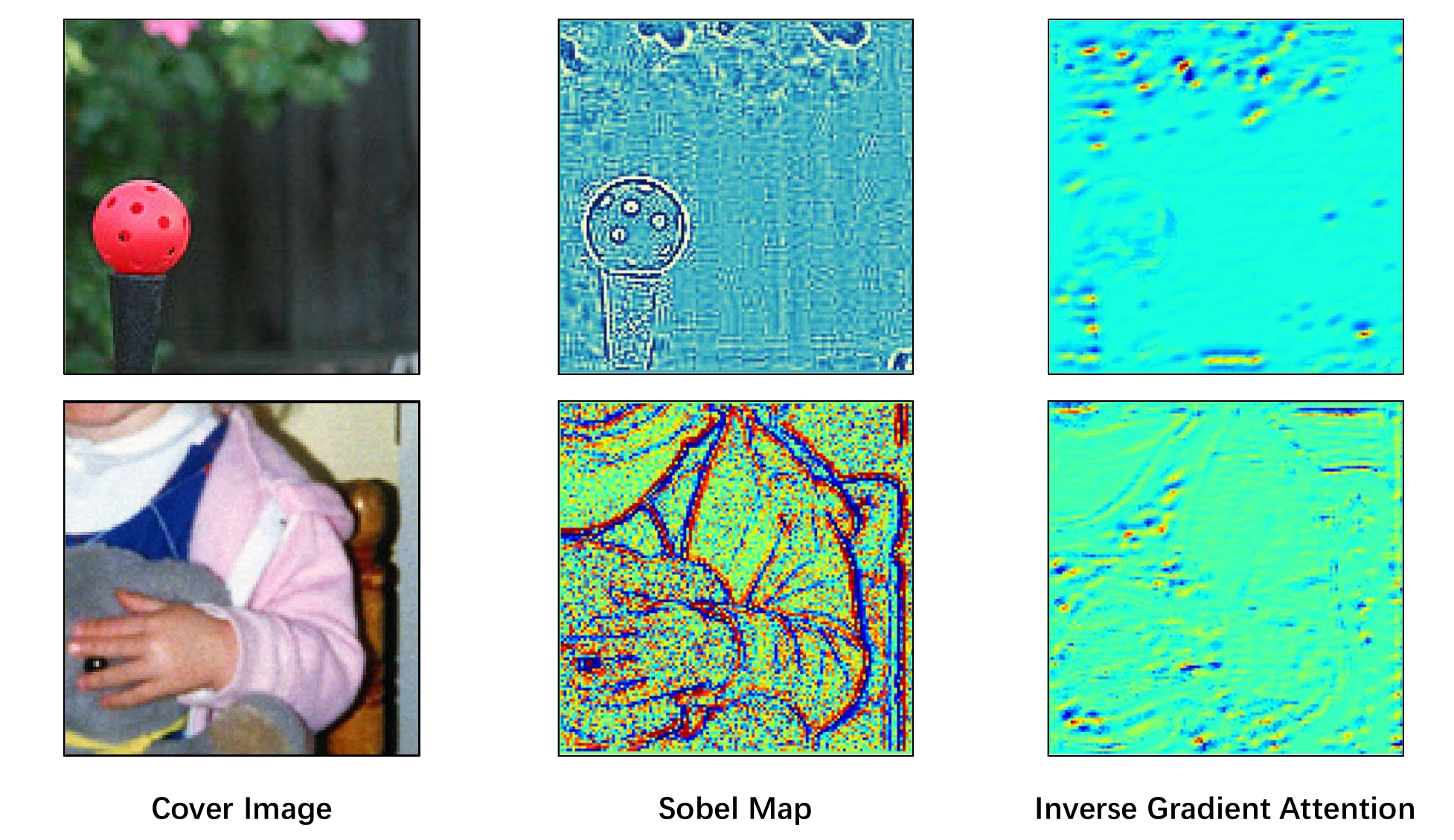}
\end{center}
\caption{The cover image, the sobel map and the corresponding inverse gradient attention heatmap.
}
\label{fig:sobel_iga_image}
\end{figure}

In this work, we adopt the Sobel operation as the high-frequency area extractor. Because it is simple to implement but efficient to stand out edge regions, which are the high-frequency areas in the image frequency domain. Through the visualization of the inverse gradient attention mask, we perceive that some edge regions are activated strongly, resonant with the edge map acquired from Sobel operation.
Fig.~\ref{fig:sobel_iga_image} shows the corresponding Sobel Map and Inverse Gradient Attention over cover images. Some similarities of the activated high frequency areas between the sobel map and inverse gradient attention are shared.

\begin{figure}[t]
\begin{center}
\includegraphics[scale=0.52]{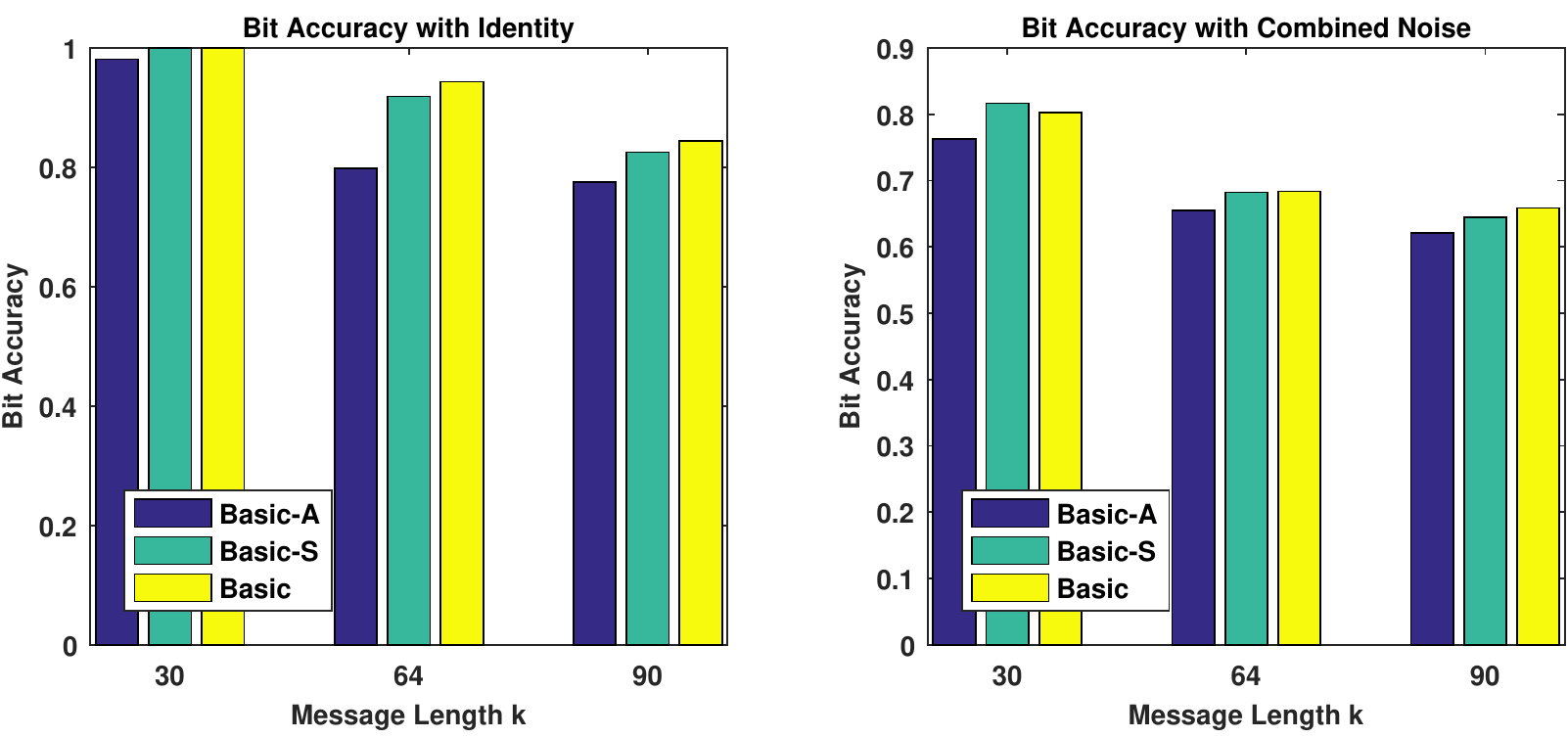}
\end{center}
\caption{The bit-wise accuracy of encoded message performance comparisons of the model equipped with proposed inverse gradient attention or the Sobel operator on COCO dataset. Model-S represents a model using Sobel edge operator and Model-A represents a model equipped with the proposed inverse gradient attention. Basic represents the basic Encoder-Decoder framework without introducing any high-frequency area extractor.}
\label{fig:bpa_coco}
\end{figure}

\begin{figure}[t]
\begin{center}
\includegraphics[scale=0.52]{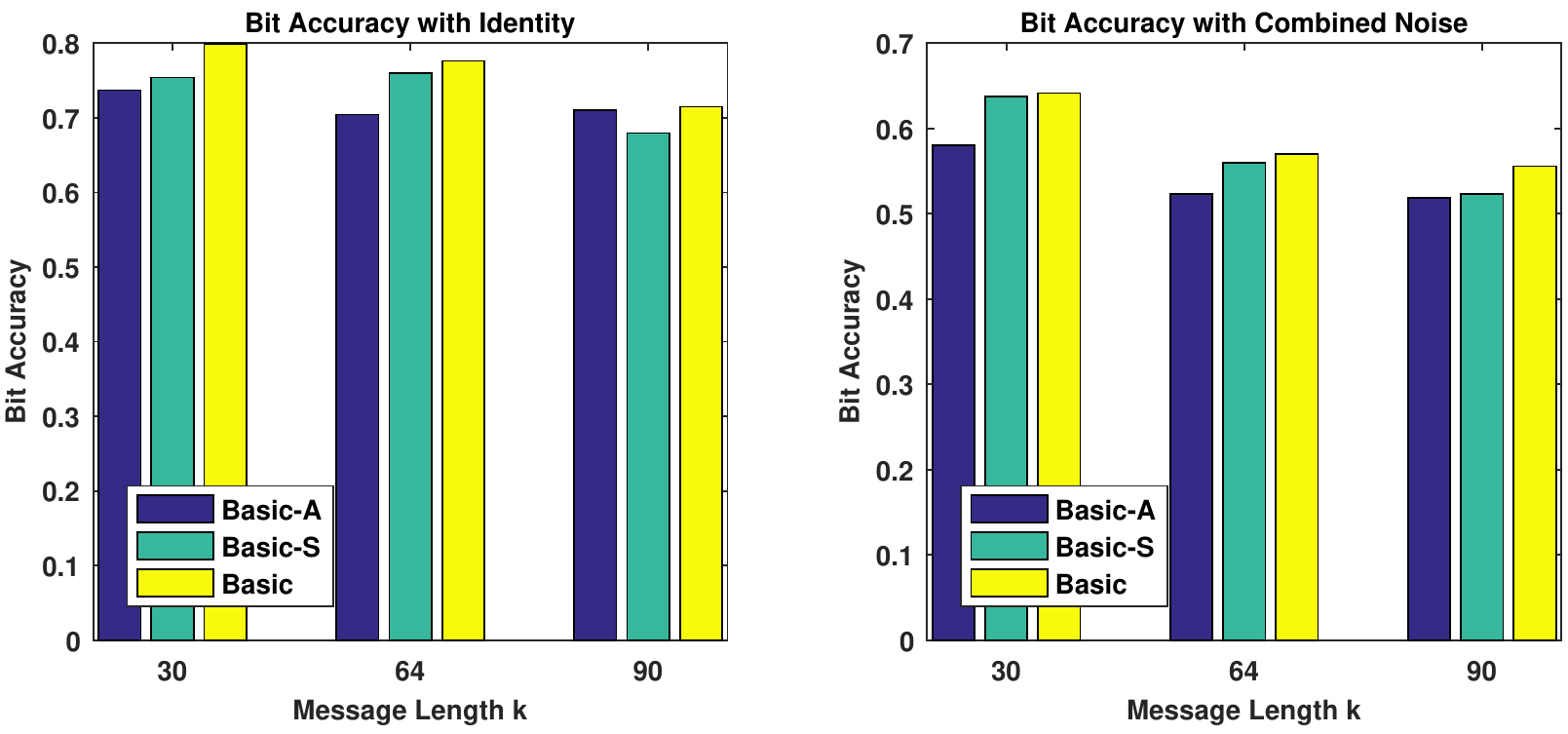}
\end{center}
\caption{The bit-wise accuracy of encoded message performance comparison of the model equipped with proposed inverse gradient attention or the Sobel operator on DIV2K dataset.}
\label{fig:bpa_div2k}
\end{figure}

According to the observation, we further conduct experiments by substituting the inverse gradient attention module with the Sobel operator in our data hiding framework. The comparative experimental results on COCO and DIV2K datasets are shown in Fig.~\ref{fig:bpa_coco} and Fig.~\ref{fig:bpa_div2k}. It can be seen that by adopting the Sobel operator for data hiding, the model also achieves promising performance, which further verifies our idea. The empirical result shows attending on pixels with rapidly changing frequencies (\textit{i.e.}, edge regions) generally has a similar effect as inverse gradient attention for model robustness enhancement. However, we can further discover that the proposed IGA model still receives better results
in all cases than the model with Sobel map. It is because the proposed IGA mechanism is able to attend on pixels adaptively toward message reconstruction objectives. It also indicates not all edge regions are suitable for information hiding.

Moreover, our empirical observations are supported by relevant inspiring works~\cite{lf_proof,wang2020high}. Specifically, the proposed proposition in ~\cite{lf_proof} proves the input perturbation is bounded by the low-frequencies. Thus, the high-frequency perturbations do not lead to large visual distortions. Besides, the human vision system is more sensitive to low-frequency components, while deep neural network models have high-frequency preferences for decision marking proved by~\cite{wang2020high}. Together with the aforementioned empirical verification, we can safely state locating robust pixels in high-frequency regions of images as IGA does is more suitable for data hiding compared to low-frequency areas.

\section{Conclusion}\label{sec:conc}

In the paper, we propose a novel end-to-end deep data hiding model with Inverse Gradient Attention (IGA) mechanism allowing the model to spotlight robust pixels and assign larger weights to them for data hiding. The model equipped with the proposed IGA module is able to adaptively and robustly embed more desired data. Besides, we adopt a symmetric message coding module to map the binary message recovery onto a low dimensional space with real values. It further improves the capacity and robustness of the model. From extensive experimental results, our proposed IGA model is able to achieve superior performance than current mainstream data hiding models. Moreover, we further discuss the strong connections between the proposed inverse gradient attention with high-frequency regions within images.

\bibliography{iga_aaai}
\bibliographystyle{aaai22}

\end{document}